\documentclass[10pt,twocolumn]{article} 
\usepackage{simpleConference}
\usepackage{times,ulem}
\usepackage{graphicx}
\usepackage{amssymb}
\usepackage{url,hyperref}
\usepackage{graphicx}
\usepackage{caption}
\usepackage{makecell}
\usepackage{booktabs}
\usepackage{mathtools}

\usepackage{float}
\usepackage{array}
\usepackage{multirow}
\usepackage{tabularx}
\usepackage{colortbl}
\usepackage{hyperref}
\usepackage{hhline}
\usepackage{array}
\usepackage{marvosym}
\usepackage{pgf}
\usepackage{hyperref}
\usepackage{indentfirst}
\usepackage{amsmath}

\usepackage{authblk}

\begin{document}

\title{Exploring Motion Boundaries in an End-to-End Network for Vision-based Parkinson's Severity Assessment}

\author{{Amirhossein Dadashzadeh$^{1}$}, {Alan Whone$^{2}$} , {Michal Rolinski$^{2}$}, {Majid Mirmehdi$^{1}$}}

\affil{ \it{Department of Computer Science, University of Bristol, Bristol, UK}$^{1}$ \\
{\it{{Department of Neurology, Southmead Hospital, Bristol, UK}}$^{2}$ }\\
{\it{Translational Health Sciences, University of Bristol, Bristol, UK}$^{2}$}\\
{\it{\normalsize {\{a.dadashzadeh, alan.whone, michal.rolinski\}@bristol.ac.uk}, majid@cs.bris.ac.uk}}}




\maketitle
\thispagestyle{empty}

\begin{abstract}
\it{Evaluating neurological disorders such as Parkinson’s disease (PD) is a challenging task that requires the assessment of several motor and non-motor functions. In this paper, we present an end-to-end deep learning framework to measure PD severity in two important components,  hand movement and gait, of the Unified Parkinson’s Disease Rating Scale (UPDRS). Our method leverages on an Inflated 3D CNN trained by a temporal segment framework to learn spatial and long temporal structure in video data. We also deploy a temporal attention mechanism to boost the performance of our model. Further, motion boundaries are explored as an extra input modality to assist in obfuscating the effects of camera motion for better movement assessment. We ablate the effects of different data modalities on the accuracy of the proposed network and compare with other popular architectures. We evaluate our proposed method on a dataset of 25  PD patients,  obtaining  72.3\% and 77.1\% top-1 accuracy on hand movement and gait tasks respectively.}

\end{abstract}

\maketitle

\section{Introduction}
\label{sec:introduction}

Parkinson’s disease (PD) is the second most common neurodegenerative disorder after Alzheimer’s dementia~\cite{sama2012dyskinesia}. {The characteristic motor features include slowness of movement (bradykinesia), stiffness (rigidity), tremor and postural instability}~\cite{zhao2008factors}. These symptoms affect patients {in performing everyday tasks} and impact the quality of their life.
Regular clinical assessment and close monitoring of the signs and symptoms of PD are required to tailor symptomatic treatments and optimize disease control. Further, accurate quantification of disease progression is vital in the trials of any drugs or interventions  that are designed to improve or modify the course of the {condition}.

Assessment of motor symptoms in PD patients is usually performed in clinical settings to evaluate the degree of rigidity and bradykinesia. Typically, the patient is asked to perform {an elaborate series} of specific physical tasks -- such as {opening and closing their hand in a rapid succession,} {i.e. gripping and letting go}, walking at usual pace for several metres. and so on, {whilst being assessed by a  
{PD {physician}  or specifically-trained nurse who {makes an evaluation}.} In formal settings, such as drug trials or research studies, the {clinical} assessment is usually scored, employing a global recognized scale, the Unified Parkinson's Disease Rating Scale (UPDRS)} \cite{updrs2008movement}, where motor evaluation consists of 33 separate examiner-defined tests. Generally, clinicians quantify the severity of each action with a score ranging {from} 0 (normal) to 4 (most severe). 
However, such a process of assessment and scoring is highly subjective and {necessitates the} {expense of an available rater}.  Therefore, automating the PD assessment process may assist in eliminating these shortcomings \cite{cunningham2012computer}.

\begin{figure*}[h]
\centering
\includegraphics[scale=0.16]{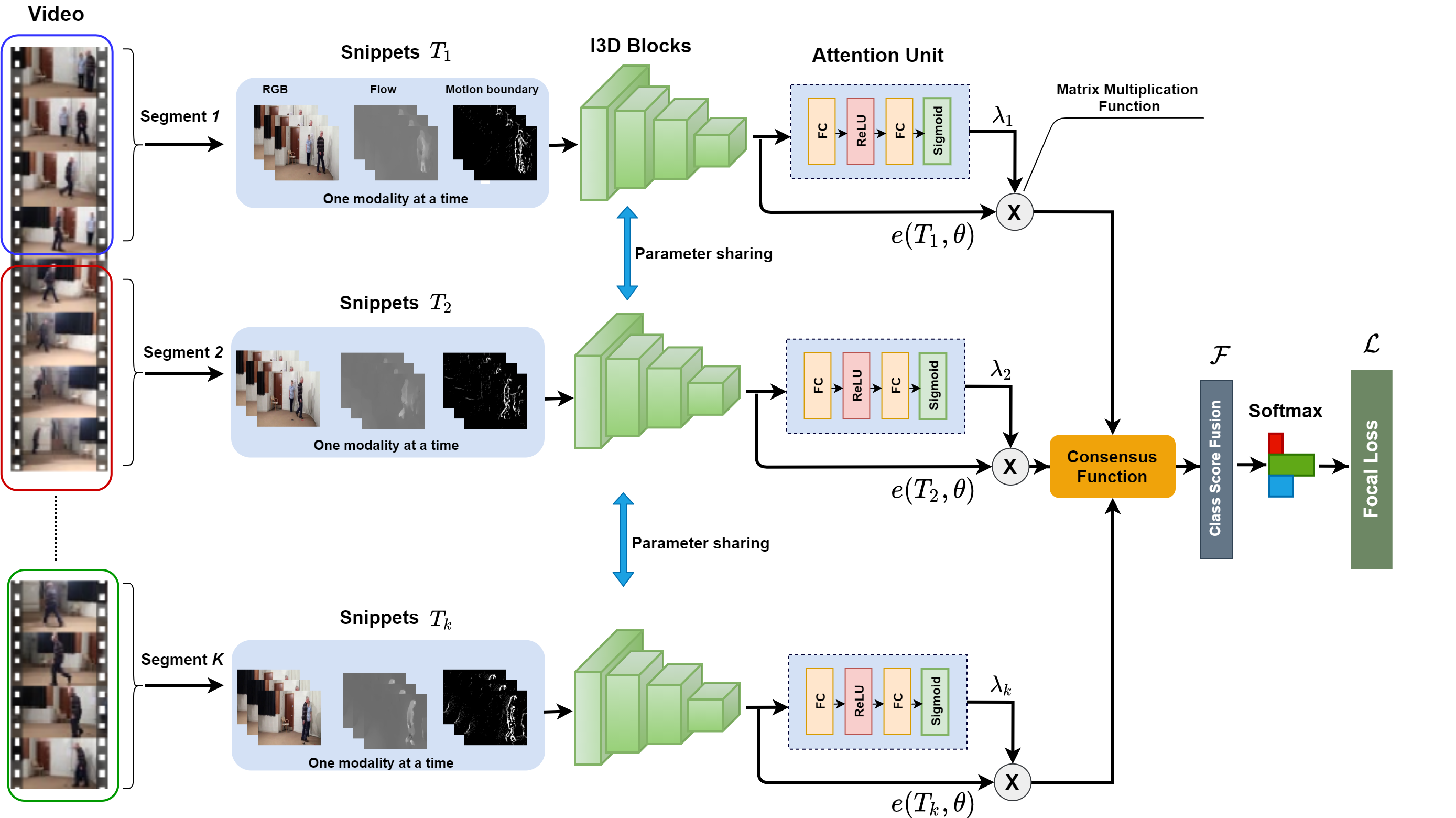}
\caption{Architecture of the proposed method for PD severity assessment task. The whole model can be trained in an end-to-end manner by only one loss function. The main steps are as follows: (i) Extracting spatial and temporal feature representations from $K$ video snippets using a single I3D network that shares all of its weights with the other branches. (ii) Computing an attention weight for each video snippet by an attention unit. 
{(iii) Weighting each feature vector by its corresponding attention weight before being forwarded to the consensus function}, (iv) Using a Softmax layer to output class score predictions. Note that at every training and testing process, the network takes one input modality amongst RGB, optical flow and motion boundaries.}
\label{fig:tsn-net} 
\end{figure*}

{In this paper, we present a simple, novel, end-to-end approach to evaluate the}
severity of PD motor state in clinical neuroscientific studies from {only} video data, based on the UPDRS scale. To suppress the influence of {arbitrary} camera motions, commonly found in real-world video data,
we deploy motion boundary features \cite{dalal2006human} computed {via} optical flow. We use such features along with other input modalities (i.e. RGB and optical flow) in a multi-stream, {deep learning} configuration to enhance the robustness of our model. We {adapt the I3D CNN  (Convolutional Neural Network)} \cite{carreira2017quo} to directly learn  {spatial and temporal features from RGB, Flow, and Motion Boundaries} 
at a low computational cost. {We also  {model} long-range temporal structure in the patient's action since {assessing} only a few moments {of} an action could result in different scores by a {rater}, e.g. {{rapid} hand opening and closing sequences} may be very similar {in part}, but in one case the hand may fail to keep up {a} consistent {amplitude and speed of} movement towards the end of the sequence due to fatiguing (as occurs in PD) or may start badly at the start of the sequence but get better as the action evolves.}
To this end, we adopt a sparse temporal sampling strategy (as proposed in \cite{wang2016temporal}) to train our network.
{This allows for} stacks of a few consecutive frames from different segments of the input video to be processed by the 3D CNN independently at inference time and their final scores averaged only at the end (see Figure \ref{fig:tsn-net}). 

Inspired by the success of `attention', now commonly used in deep learning networks, e.g. for human action recognition \cite{pei2017temporal,long2018attention}, we engage attention units which assign individual attention weights over each feature vector. This enables a more reliable model as it allows our network to focus more heavily on the critical segments of a video which may contain  absolute classification information. This is motivated by the fact that, in some cases, clinicians are able to pass judgement on a patient based on momentary actions, e.g. an interruption or hesitation during the hand movement task.

We use a dataset collected from 25 {clinically diagnosed} PD patients who underwent UPDRS assessments of their motor function after withholding symptom improving dopaminergic medication overnight, focusing on the rapid hand opening and closing and gait components. 
We train and test our model via a {subject-level N-fold cross validation scheme} to {evaluate its} performance and compare against other popular deep learning {architectures} -- {in particular  to demonstrate the importance of the use of motion boundaries.} 
To the best of our knowledge, ours is the first work to propose an end-to-end deep learning framework for automatic PD severity assessment based on UPDRS scores from non-skeleton-based data.

In summary, our main contributions are as follows: (i) we leverage recent advances from deep learning techniques in human action recognition and combine them with a temporal attention-based approach to find a practical design choice for video-based PD severity prediction, (ii) in order to reduce the camera motion effect and increase accuracy, we propose to use motion boundaries as an extra input in our multi-stream configuration,
(iii) we quantitatively compare different architectures and different input modalities, and include ablation studies to determine the influence of attention and each modality for two PD motor function tasks.

Next, in Section \ref{sec:Related-works}, we describe related works briefly. In Section \ref{sec:Proposed-Approach}, our proposed methodology is outlined. Experiments and comparative results are presented in Section \ref{sec:results}, and finally we conclude the paper in Section \ref{sec::conclusions}.

\section{Related works}
\label{sec:Related-works}

Recently, challenges in PD have been addressed through machine learning techniques, mostly {employing}  wearable sensors, such as \cite{jeon2017automatic,pereira2019survey,abdulhay2018gait,hobert2019progressive}.
For example, \cite{jeon2017automatic} performed a comparative study of various machine learning algorithms, such as decision trees, support vector machines, discriminant analysis, random forests, and k-nearest-neighbor on data  from a wrist-worn  wearable device to classify hand tremor severity. Evaluated on 85 patients, the highest accuracy obtained was 85.6\% by a decision tree classifier.

{Studies that have applied deep learning techniques, such as CNNs and RNNs (Recurrent Neural Networks), to PD severity automatically from wearable sensor data include \cite{zhao2018hybrid,xia2019dual,el2020deep,hssayeni2019wearable,sigcha2020deep}}. 
For example, \cite{zhao2018hybrid} developed a two-stream deep learning architecture, 
including a 5-layer CNN and a 2-layer LSTM (Long Short-Term Memory) network to capture the spatial and temporal features of gait data.
{Their model was trained and tested on three public Vertical Ground Reaction Force datasets collected by foot sensors from 93 patients with idiopathic PD and 73 healthy controls. A maximum accuracy of 98.8\% was reported for classification of PD patients with different severity. }

{However,} the use of force sensors, wearables and other on-body sensors have many limitations. They are mostly inconvenient, and sometimes very difficult, to attach to patients, but more importantly, they do not provide the spectrum of rich information a visual sensor can present. Moreover, a camera system in the clinic is passive and not so intrusive as an attached sensor that can {produce} discomfort and unease.

{With the rise and wide availability of depth-based sensors (e.g. the Kinect), which reduces the burden of capturing 3D joints, they have been adopted in {many} health-related applications, such as \cite{hall2016designing,li2018classification,khokhlova2019normal}.} {For example,  \cite{khokhlova2019normal}
{used Kinect V2 skeletons to collect a dataset of normal and pathological gait examples from 27 subjects. Shoe-sole padding was placed into the right shoe of each person to simulate gait problems. They obtained dynamic features of lower limbs in their dataset to analyse the symmetry of gait and then applied these features in an LSTM-based model to learn the difference between normal and abnormal gait. An average accuracy of $78.5\%$ was achieved by cross-validation on 10 different validation partitions.}}

{We are interested in assessing PD severity {with recourse to only RGB (and RGB-derived)} data which is an easily available modality.
Works such as \cite{li2018vision,pintea2018hand,chang2019improving}, generate RGB video features from deep learning-based pose estimation networks. 
For example,  \cite{pintea2018hand} estimated the frequency of hand tremors in Parkinson’s patients by applying Convolutional Pose Machines (CPM)  \cite{wei2016convolutional} and Kalman filtering to detect and track hand motion, and then subtracting the original hand locations from their smooth trajectory to estimate the tremor frequency. 
\cite{li2018vision} used movement trajectory characteristics (e.g. kinematic, frequency) extracted using CPM  to train random forests to assess the severity of Parkinsonism in leg agility and toe tapping tasks.} {\cite{chang2019improving}  used OpenPose \cite{cao2018openpose} to extract frame-level feature keypoints from finger taps and hand tremor tasks. They encoded these into a single task-level feature vector by using 15 statistical functions, such as max, min, mean, median, standard deviation, Fisher vectors, and so on. This single task-level feature vector was then fed into a feedforward neural  network to classify each subject into normal and abnormal classes for each hand at the individual task level.}
The key limitations of these pose-based methods are that they rely on hand-crafted feature extraction and pre-processing steps which limits their models' representational capability.

{To obtain good visual representations, especially rich in temporal features, is a challenging task and has received considerable attention in recent years \cite{simonyan2014two,wang2016temporal,carreira2017quo,feichtenhofer2019slowfast}. In an early work, \cite{simonyan2014two} designed a two-stream CNN to model spatial and temporal features
from RGB and optical flow, respectively. These were then fused to generate the final classification scores. The Temporal Segment Network of \cite{wang2016temporal} further improved the results of the two-stream CNN at a reasonable computational cost by modelling long-range temporal structures and proposing a temporal sampling strategy for training CNNs on video data. By inflating the ImageNet pre-trained 2D kernels into 3D,  \cite{carreira2017quo} directly learned spatial and temporal features from RGB, significantly enhancing the state of the art on action recognition at the time. Their Inflated 3D ConvNet (I3D) remains a popular network of choice. Feichtenhofer et al. introduced the SlowFast architecture  \cite{feichtenhofer2019slowfast}, which uses a slow pathway to capture spatial content of a video, and a fast pathway to capture motion at fine temporal resolution. Our proposed multi-stream architecture applies an I3D backbone, based on a temporal segment strategy, and we will also consider the SlowFast approach for our comparative analysis.}


\section{Proposed Approach}
\label{sec:Proposed-Approach}

Our aim is to learn an end-to-end, deep learning model for movement disorder severity assessment in Parkinson's patients, {without resort to joint data or elaborate annotations}. Given a video from a patient {in the clinic} performing a UPDRS test task, such as hand {opening and closing}, our model exploits the motion information in the scene to predict a score depending on how well the task was carried out. {Our only annotation is the UPDRS score for the test, as determined by an expert clinical {neuroscience rater}.}
Figure \ref{fig:tsn-net} illustrates an overview of our network and approach. In the following, we explain the details of our method and its training procedure.

\begin{figure*}[h]
\centerline{\includegraphics[scale=0.2]{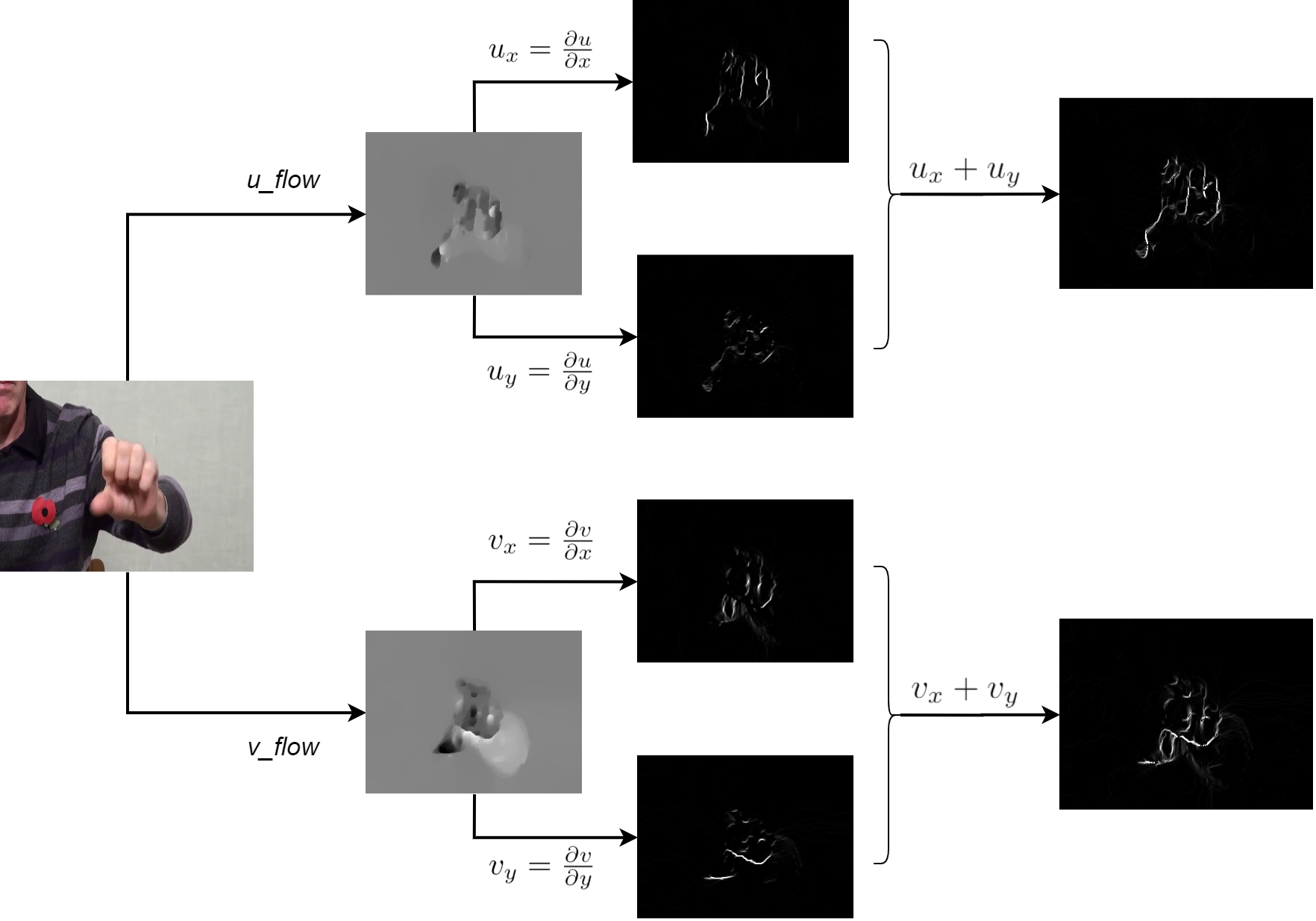}}
\caption{{Motion boundary computation from optical flow components $u$ and $v$. For each flow component, we compute two motion boundaries via derivatives for the horizontal and vertical flow components. Then the final motion boundaries are obtained by their sum. It is clear that optical flow contains constant motion in the background which is removed after computing motion boundaries.}}
\label{fig:motion-b}
\end{figure*}

{\bf Network architecture -- } Following \cite{wang2016temporal},
we use {sparse temporal sampling  for our model training.}
As shown in Figure \ref{fig:tsn-net}, 
we {first} split the video into $K$ segments, each of which is randomly sampled into a short snippet to form sparse sampling of the whole clip into $K$ snippets {$\{T_i, i=1..K\}$ -- {with each snippet generated in three formats (RGB, Flow, and Motion Boundaries, but not individually specified here for reasons of simplicity and brevity)}}. {Then, similar to \cite{liu2018t}, we apply a 3D CNN as the backbone of this {framework} {to directly learn spatial and temporal features from video snippets.}}
To overcome the increased parameter space and associated risks of overfitting resulting from this change,  I3D \cite{carreira2017quo} is deployed {as the 3D CNN} which inflates all the 2D convolution filters used by the Inception V1 architecture \cite{szegedy2015going} into 3D convolutions allowing a deep 3D ConvNet with many fewer parameters.  We found this strategy very efficient to analyse the complex motions of our video data which are by nature relatively long.

The spatial and temporal feature maps of the last convolutional layer of I3D for each video snippet feed into an attention unit that consists of two fully connected (FC) layers interspersed  by a ReLU activation function and a Sigmoid function to generate attention weights $\lambda$ ($0.0 \leq \lambda \leq 1.0$) for each video snippet. 
{This is based on the attention module proposed in \cite{nguyen2018weakly}}.

{Then,} in the forward pass of the system, the encoded, attention-weighted features are used to modulate the global average pooling and therefore compiled via the consensus function {$C(.)$}  {to produce class score fusion {$\mathcal{F}$}} {of length $M$} over $K$ video snippets,  
 \begin{equation} 
 \mathcal{F} = C(.) = \frac{\Sigma_{i=1}^{K}(\lambda_i e(T_i,\theta))}{K} ,
 \label{eq:fusion}
\end{equation}
{where $e(.)$ is the encoding function
and {$\theta$ are the network parameters.}
A Softmax on $\mathcal{F}$ then provides the probability distribution {$p$} of the UPDRS class scores of the video clip, i.e.}
\begin{equation} 
p = \frac{\exp \mathcal{F}{_i}}{\Sigma_{j=1}^{X} \exp \mathcal{F}_j} ~.    
 \label{eq:softmax}
\end{equation}
{\bf Motion Boundaries -- }
{Previous works, such as  \cite{simonyan2014two,wang2016temporal,carreira2017quo}, have shown the importance of using optical flow in deep learning-based human action recognition. However, optical flow fields represent the absolute motion, making the disentanglement of object-level  and camera motions a significant challenge \cite{Chapel2020moving}.} 
{\cite{wang2016temporal} proposed to use warped flow \cite{wang2013action} to cancel out the camera motion. However, warped flow did not results in a better performance than normal optical flow in their work. Moreover,   computing this modality can be computationally very expensive \cite{wang2013action}.} 

To address this problem, we need a new input stream that better encodes the relative motion between pixels. Thus, we use motion boundaries, initially proposed in the context of human detection \cite{dalal2006human},
to remove constant motion and therefore suppress the influence of camera motions.

{In a similar fashion to \cite{dalal2006human}, we compute} motion boundaries {simply} by a derivative operation on the optical flow components, as shown in Figure \ref{fig:motion-b}. Formally, let $u_x= \frac{\partial u}{\partial x}$ and $u_y= \frac{\partial u}{\partial y}$ represent the $x$ and $y$ derivatives of horizontal optical flow, and $v_x= \frac{\partial v}{\partial x}$ and $v_y= \frac{\partial v}{\partial y}$ represent the $x$ and $y$ derivatives of vertical optical flow respectively. Then, for any frame $j$, 
\begin{equation} 
B_u^j=f(u_x^j, u_y^j),
B_v^j=f(v_x^j, v_y^j),
\end{equation}
\noindent where $B_u$ represents the motion boundary in horizontal optical flow $u$, and $B_v$ represents the motion boundary in vertical optical flow $v$, and $f$ is {a summing} function. It is clear that, for a video clip with $N$ frames, $(N-1)*2$ motion boundary frames  are computed.

{\bf Class imbalance --}
{In the PD dataset used in this study (see details in Section \ref{sec:results}), the number of videos belonging to UPDRS scores 3 and 4 is significantly lower} 
{than those belonging to the other classes. 
Therefore, we have a class imbalance problem which can lead to a model biased towards}
{the classes with large number of samples.}

In order to mitigate this problem, we apply two strategies. In the first, we group the scores into three classes: score 0 for normal subjects - i.e. patients who are at very early stage of PD {and may still have one unaffected upper limb}, score 1-2 for subjects with mild symptoms, and score 3-4 for  subjects with severe symptoms. In the second strategy, we utilize an extended version of the normal class entropy loss, called focal loss \cite{lin2017focal}, to train our multi-class classification task. The original focal loss was proposed for single-class object detection in order to down-weight easy classes and better weight rarer classes by adding a factor to the standard cross entropy loss.
Then, our loss function can be stated as
\begin{equation} 
\mathcal{L}(y,p)= -\alpha (1-p_{})^\gamma y_{} \log p_{} ~,
\label{eq:loss}
\end{equation}
\noindent where $y$ is the UPDRS groundtruth label and  $\gamma$ adjusts the rate at which easy samples are down-weighted. This adds a modulating factor $\alpha (1-p_{})^\gamma$ to the cross-entropy loss. When  a sample is classified with a high probability, i.e. $p_{}$ is large, the value of the modulating factor is small and the loss for that sample is down-weighted. In contrast, when a  hard sample is misclassified with low probability, the modulating factor is large, increasing that sample’s contributions to the total loss. The value of $\alpha$ is  prefixed ($0 < \alpha \leq 1$) to balance the importance of samples belonging to different classes.  When $\alpha=1$ and $\gamma=0$, the focal loss is
equivalent to cross-entropy loss.


\section{Experiments} \label{sec:results}
In this section, we first present our PD dataset used for evaluating our proposed method. Then, we provide the experimental setup and our detailed ablation study of various aspects of our model. 
Finally, we compare our model with the state of the art works in human action recognition.

\begin{figure*}[h]

\centerline{\includegraphics[scale=0.28]{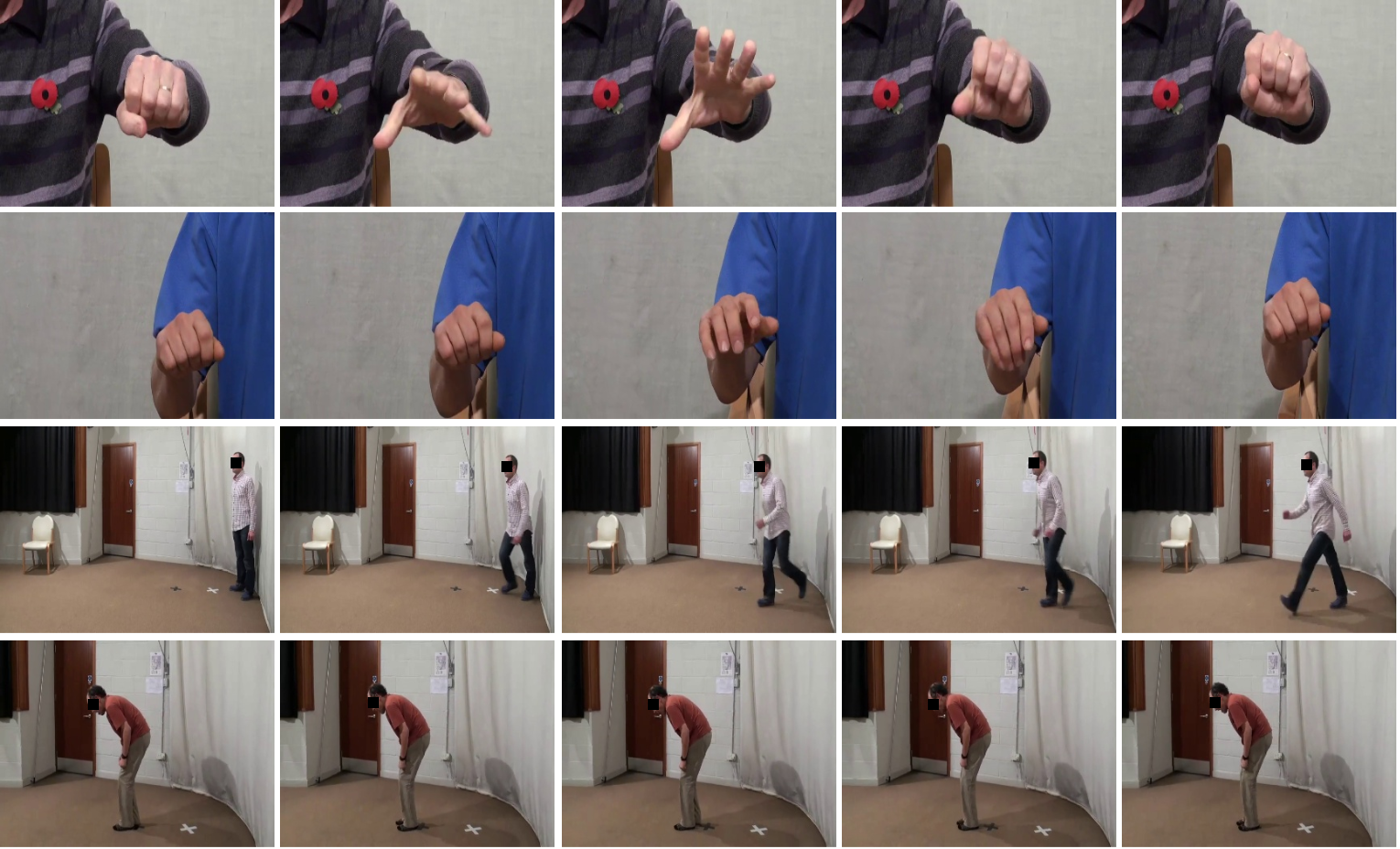}}
\caption{Sample frames from different patients at varying severity levels (top two for hand movement, lower two for gait).}
\label{fig:sample}
\end{figure*}

\begin{figure}[h]

\centerline{\includegraphics[scale=0.27]{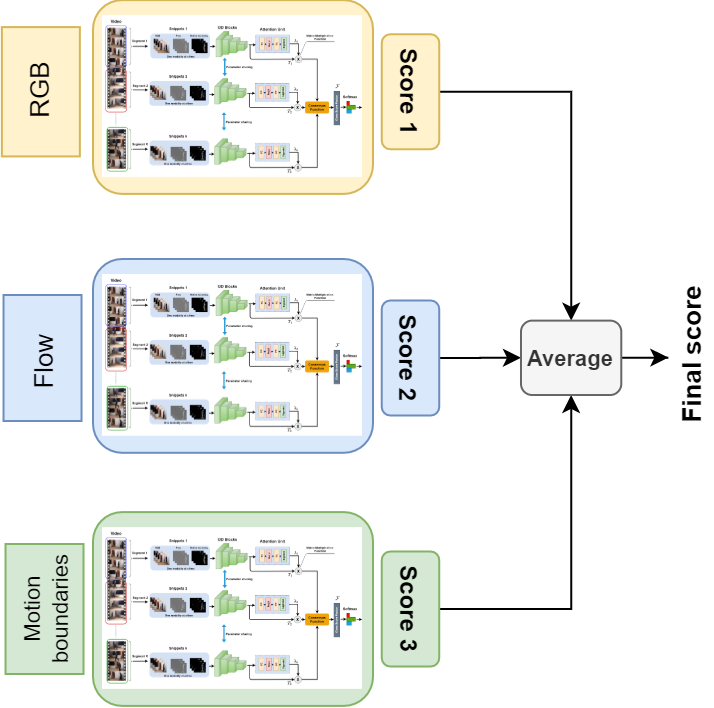}}
\caption{An overview of our multi-stream configuration. We train our model with each different input modality separately and then use a late fusion approach at test time to average over all predicted scores.}
\label{fig:multi-s}
\end{figure}

\begin{table} 
\scriptsize
\centering
\caption {Details of our PD dataset.}
\resizebox{0.5\textwidth}{!}{%
\begin{tabular}{c|c|c||c|c|l}
\cline{2-5}
\multicolumn{1}{l|}{}  & \multicolumn{2}{c||}{\textbf{Hand movement}} & \multicolumn{2}{c|}{\textbf{Gait}}    &  \\ \cline{1-5}
\multicolumn{1}{|l|}{\textbf{Score}} & \textbf{\#video} & \textbf{\begin{tabular}[c]{@{}c@{}}\#frame\\ min/max\end{tabular}} & \textbf{\#video} & \textbf{\begin{tabular}[c]{@{}c@{}}\#frame\\ min/max\end{tabular}} &  \\ \cline{1-5} 
\multicolumn{1}{|c|}{\textbf{Normal (0)}}   &  180   & 131/312   & 171  & 473/980   &  \\ \cline{1-5}
\multicolumn{1}{|c|}{\textbf{Mild (1-2)}}   &  500  & 123/717   & 180  & 580/5007  &  \\ \cline{1-5}
\multicolumn{1}{|c|}{\textbf{Severe (3-4)}} &  24  & 202/1210   & 3   & 1367/3012   &  \\ \cline{1-5}

\end{tabular}}
\label{Tabel::dataset}
\end{table}

\subsection{Dataset}

The Parkinson's Disease dataset used in this study
contains video data from 25 PD patients {tested longitudinally at 8 week intervals over time. Subjects were} between the ages of 41 to 72 years {and performed} UPDRS tasks and their scores {were} assigned by {trained clinical raters}. Videos were captured at 25fps at a resolution of 1920$\times$1080,  using a single RGB camera (SONY HXR-NX3). 
{Our dataset consists of 1058 videos spanning two different UPDRS tasks: hand movement and gait.} 
In the first task, the patients had to open and close their hand (each hand separately) 10  times, as fully and as quickly as possible. The second task is gait analysis in which the patients walked 10 metres at a comfortable pace and then returned to their starting point. Table \ref{Tabel::dataset} shows the number of videos in each of our score classes, as well as their minimum/maximum number of frames for each UPDRS task. Figure \ref{fig:sample} shows sample frames from  our dataset, selected from four subjects with different PD severity levels  performing hand movement and gait tasks.

{\bf{Implementation details --}} 
The input videos were reduced to a resolution of $340\times256$ pixels. We used Pytorch to implement our  models and TV-L1 \cite{zach2007duality} for computing optical flow fields. The focal loss (Eq. \ref{eq:loss})  parameters were set to $\alpha= 0.5$ and $\gamma=2$ for all
experiments. We applied Adam optimization with a learning rate of $0.00001$, and batch size $2$ to optimize our model parameters. Dropout was applied with a ratio of 0.7 before the output layer of our I3D network. All models were trained for 120 epochs using one Nvidia RTX 2048TI GPU under Cuda 10.1 with cuDNN 7.6.

\subsection{Experimental Setup}

{\bf{Training and Testing details --}} Each video was split into $K=4$ equal segments along the temporal axis. {Preserving chronological order,} we randomly sampled $32$  frames within each video segment as a snippet.  
{The length of our snippets is relatively larger than the length of snippets used 
in \cite{wang2016temporal}. We verified empirically that for our PD task sampling these larger snippets can provide more application-specific motion characteristics  to our network.}

Since in the training step all I3D models share their parameters, our trained model behaves like the original I3D network \cite{carreira2017quo} during testing. Therefore, we did not use temporal sampling when testing our model, allowing us to draw fair comparison with other models who also tested without temporal sampling, such as \cite{simonyan2014two,carreira2017quo,feichtenhofer2019slowfast}. Specifically, during testing we used 64 {non-sampled snippets} per video, each containing $16$ consecutive frames. The prediction scores of all these snippets were then averaged across each or combined modalities to get a video-level score {(as illustrated in Figure \ref{fig:multi-s}). Note, this follows the same approach as \cite{carreira2017quo} where RGB and Flow were averaged at test time.}

To avoid overfitting, we initialised our network by pretraining on Kinetics \cite{kay2017kinetics}, and applied augmentation for all frames within each training snippet, including  scale jittering, corner cropping, and horizontal flipping. 

{\bf Evaluation metrics --} 
We used 5-fold cross validation for 5 batches (given our 25 patients) to yield unbiased performance of the models and report the  $F_1$ score over the average validation scores.

\begin{table*}
\footnotesize
\centering
\caption{$F_1$ score results of our proposed network for both hand movement and gait tasks with different input modalities, with and without attention units. The last column shows the average results across both tasks. { All results are given in \%.} \label{Table:data-augmentation}}
\begin{tabular}{l|c|c||c|c||c|c||}
\cline{2-7}
\textbf{}    & \multicolumn{2}{c||}{\textbf{Hand Movement}}    & \multicolumn{2}{c||}{\textbf{Gait}}    & \multicolumn{2}{c||}{\textbf{Average}} \\ \hline
\multicolumn{1}{|l||}{{\bf Input Modalities}}     & \multicolumn{1}{c|}{$+$attention} & \multicolumn{1}{c||}{$-$attention} & \multicolumn{1}{c|}{$+$attention} & \multicolumn{1}{c||}{$-$attention} & $+$attention      & $-$attention     \\ \hline
\multicolumn{1}{|l||}{RGB}                    & 68.4                             & 65.2                              & 74.8                             & 73.7                              & 71.6             & 69.4             \\ \hline
\multicolumn{1}{|l||}{Flow}                   & 71.0                             & 68.6                              & 76.5                             & 74.1                              & 73.7             & 71.3             \\ \hline
\multicolumn{1}{|l||}{Motion Boundaries}        & \textbf{72.3}                    & 70.0                              & 76.8                             & {76.5}                     & \textbf{74.5}    & 73.2             \\ \hline

\multicolumn{1}{|l||}{RGB + Flow}             & 69.9                             & 68.8                              & 76.2                             & 75.1                              & 73.0             & 71.9            \\ \hline
\multicolumn{1}{|l||}{RGB + Motion Boundaries}  & 70.4                             & 70.1                              & 75.4                             & 72.3                              & 72.9             & 71.2             \\ \hline
\multicolumn{1}{|l||}{Flow + Motion Boundaries} & 71.7                             & {71.7}                     & \textbf{77.1}                    & 76.2                              & 74.4             & {73.9}    \\ \hline
\multicolumn{1}{|l||}{All Modalities}            & 71.1                             & 70.2                              & \textbf{77.1}                    & 75.1                              & 74.1             & 72.6             \\ \hline

\end{tabular}
\label{tabel::as}
\end{table*}

\begin{table*}
\footnotesize
\centering

\caption{Comparison of our method with different state-of-the-art architectures. {MBs is for Motion Boundaries and all results are given in \%.}} \label{Table:data-augmentation}

\begin{tabular}{|l||c|c|c||c|c||c||}
\hline \hline
\textbf{Model}   &  \textbf{RGB}&  \textbf{Flow} & \textbf{MBs} & \textbf{Hand Movement} & \textbf{Gait} & \textbf{Average} \\ \hline

Two-Stream \cite{simonyan2014two}     &&\checkmark&  \checkmark                           &  $60.3$                      & $56.7$         & $58.5$          \\ \hline

TSN \cite{wang2016temporal}         &  & \checkmark &   \checkmark                                  &  $70.1$                      & $75.7$& $72.9$      \\ \hline

I3D \cite{carreira2017quo}           && \checkmark&    \checkmark                            & $69.1$                       &  $73.1$  &  $71.1$                \\ \hline

SlowFast \cite{feichtenhofer2019slowfast}     &\checkmark&&                            & $67.1$                     &  $66.9$           & $67.0$         \\ \hline

TSN + SlowFast                                 &\checkmark   &&  & $68.4$                  &  $68.9$           & $68.6$  \\ \hline

Proposed Method  \tiny{w/o Focal loss}      & &&\checkmark&                      $70.7$                       &        $75.7$           & $73.2$  \\ \hline

Proposed Method                         &&\checkmark&\checkmark&  $71.7$                      &  \bf{ 77.1}            & $74.4$ \\ \hline 

Proposed Method                          & &&\checkmark&  \bf{72.3}                      &  $76.8$                   & \bf{74.5}\\ \hline

\end{tabular}
\label{table:comp}
\end{table*}

\subsection{Results including Ablation Study}
{\bf Choice of input modalities --}
{The results in Table \ref{tabel::as} show comparative evaluations on different input modalities. For the Hand Movement task, motion boundaries alone perform best at $72.3\%$ as they capture the characteristic motions in the task. 
This result improves over using RGB and Flow  by $\uparrow\!\!\!3.9\%$ and $\uparrow\!\!\!1.3\%$ respectively. When Motion Boundaries are combined with RGB and Flow, the results improve over using those modalities alone to $70.4\%$ and $71.7\%$ respectively. For the gait task, where there is much more pronounced dynamic movement spatiotemporaly, all modalities perform comparatively well, with Flow+Motion Boundaries achieving the best performance at $77.1\%$. On average, the use of Motion Boundaries is vindicated as a significant extra modality that can contribute to, or alone generate, improved results.  
}

{\bf Effect of Attention --}
{To study the influence of the attention units, we perform all our experiments with and without them. As seen in Table \ref{tabel::as}, in all experiments for hand movement and gait tasks, our model achieves better accuracy {\it with} the attention units. Again, even without attention units, Motion Boundaries play a significant role in improving the results over other modalities.}

{\bf Performance of Other Architectures -- }
{Table \ref{table:comp} provides the $F_1$ percentages of other architectures adapted to provide a UPDRS score for our application. We used the same data augmentation strategy with focal loss to train all models. All the network weights were initialized with pre-trained models from Kinetics-400, except for the SlowFast network, as one of the properties of this model is training from scratch without needing any pre-training. 
Although we examined the performance of these architectures for all possible input modalities, we only report here their best results, again except for the SlowFast network, as this model is only based on RGB input. Thus, for example for I3D \cite{carreira2017quo}, its best result is when using Flow and Motion Boundaries. As shown in the table, our proposed approach performs better than these popular networks for both hand movement and gait tasks}.

{\bf Effect of Focal Loss --}
{The importance of using our focal loss (Eq. \ref{eq:loss}) is also shown in Table \ref{table:comp} where the performance of our method when using a categorical cross-entropy loss results in an average drop of $\downarrow\!\!1.3\%$ compared to the full focal-loss based result of $74.5\%$.}

\section{Conclusions}
\label{sec::conclusions}
In this paper, we proposed an end-to-end network for the assessment of PD severity from videos for two UPDRS tasks: hand movement and gait. Our model builds upon an inflated 3D CNN trained by a temporal sampling strategy to exploit long-range temporal structure at low cost. We applied an attention mechanism along the temporal axis to provide learned attention weights for each video segment, allowing our model to focus more on the relevant parts of each video. We also proposed the use of motion boundaries as a viable input modality to suppress constant camera motion and showed its effect on the quality of the assessment scores quantitatively.  
We also evaluated the performance of several popular architectures for PD severity assessment.

One limitation of our approach is that it is unable to handle several UPDRS tasks in one training process in which we need to train and evaluate our model on each task separately. In future work, we {hope to handle this issue through unsupervied learning and adopting multiple loss functions with} an effective way to combine them.

\section*{Acknowledgements}
The authors sincerely thank for the kind donations to the Southmead Hospital Charity and from Caroline Belcher. Their generosity has made this research possible.

\normalem

\bibliographystyle{unsrt}



\end{document}